# The Challenges of HTR Model Training: Feedback from the Project
*Donner le gout de l'archive à l'ère numérique*

Béatrice Couture[1], Farah Verret[1], Maxime Gohier[2], Dominique Deslandres[1*]

1 Université de Montréal, Canada
2 Université du Québec à Rimouski, Canada

*Corresponding author: Dominique Deslandres: dominique.deslandres@umontreal.ca

**Abstract**
The arrival of handwriting recognition technologies offers new possibilities for research in heritage studies. However, it is now necessary to reflect on the experiences and the practices developed by research teams. Our use of the Transkribus platform since 2018 has led us to search for the most significant ways to improve the performance of our handwritten text recognition (HTR) models which are made to transcribe French handwriting dating from the 17th century. This article therefore reports on the impacts of creating transcribing protocols, using language models at full scale and determining the best way to use base models in order to help increase the performance of HTR models. Combining all of these elements can indeed increase the performance of a single model by more than 20% (reaching a Character Error Rate below 5%). This article also discusses some challenges regarding the collaborative nature of HTR platforms such as Transkribus and the way researchers can share their data generated in the process of creating or training handwritten text recognition models.

**List of abbreviations**
APAD: Atelier permanent d'analyse documentaire (Université de Montréal)
CER: Character Error Rate
CREMMA: Consortium Reconnaissance d'Écriture Manuscrite des Matériaux Anciens
HTR: Handwritten Text Recognition
OCR: Optical Character Recognition
WER: Word Error Rate

**I INTRODUCTION**
Since 2018, the team of *Donner le goût de l'archive à l'ère numérique*[1] has enthusiastically used the Transkribus platform and its handwritten text recognition (HTR) technologies. Doing so, we were able to automatize the transcription of, as well as the tagging and publishing, the data related to a vast corpus of judiciary and notarial records produced in Montreal (Canada) in the 17th century. The archival records used were particularly difficult to decipher and written by more than twenty-five individuals (see table 1). Therefore, during this five-year period, we have trained more than fifteen HTR models that can be grouped into two categories: single-hand models and general models (see table 2). Based on a set of approximately 1,500 manually transcribed pages, these models have been used to transcribe more than 60,000 pages (which represent 73% of our corpus).[2]

At the beginning of the project, little literature was available on HTR technologies and on the creation of models to guide us. To acquire our knowledge, we have been compelled to opt for a trial-and-error process, and that in an intuitive way. Currently, research on HTR technologies is quite dynamic, and several major studies have been published (Kahle, 2017; Massot, 2019; Muehlberger, 2019; Sanchez et al, 2019; Schlagdenhauffen, 2022; Colavizza, 2022; Kokaze, 2022; Nockels et al, 2022). However, they do not systematically analyze the different variables influencing the performance of HTR models and, as noted by [Nockels et al., 2022], literature on individual experiences on Transkribus, as for other platforms, is still scattered (see also in Stokes, 2020).

Thus, we believe that a reflexive feedback on our experience and the taking into account of these studies are highly relevant and can contribute not only to (1) scientific knowledge about the use of HTR technologies and on the factors influencing models' performance but, more importantly, to (2) the best practices for future researchers. Indeed, given the changes recently made and the future changes announced by Transkribus[3], it seems even more relevant now to initiate a common reflection on the impact of the different practices surrounding the training and sharing of models. Such a reflexive approach is essential to ensure that the work currently being done around the world on the training of HTR models is as useful as possible, regardless of the platform used, the structure of various algorithms, and the technological developments to come.

In order to achieve this double objective, we briefly present the project and corpus from which the empirical experimentation was conducted, as well as the methodology we used for training our models and analyzing their performances. Then, we discuss the specific impact of transcription protocols, language models, and base models on HTR models' performances.

---

[1] Dominique Deslandres (Université de Montréal) is the leading investigator of the project Giving the Taste of the Archive in the Digital Age: Production and Sharing of Historical and Archaeological Data on the Peoples of Montreal in the 17th Century, which was financed by an SSHRC Partnership Development Grant (2021-2024). Her co-researchers are Maxime Gohier (Université du Québec à Rimouski) and Léon Robichaud (Université de Sherbrooke). The project has multiple institutional partnerships at the Université de Montréal (Department of History; Document Management and Archives Division [DGDA]; Library of Rare Books and Special Collections; Department of Computing and Operations Research; Research Program in Historical Demography [PRDH]) as well as external partners (National Library and Archives of Quebec [BAnQ]; Pointe-à-Callière Museum, Montreal Archaeology and History Complex; Archiv-Histo Historical Society; READ-COOP).
[2] The remaining pages have still not been transcribed since it is necessary to create another ten models for less important notaries, i.e., notaries who have written between 200 and 2000 pages.
[3] In November 2022, the HTR+ algorithm developed by the CITlab of the University of Rostock was discontinued in favour of the PyLaia technology developed by the Technical University Valencia. Future changes, notably concerning Transformer neuronal systems (Ströbel, 2022), have also been announced by the READ-COOP team at the Transkribus User Conference 2022.

Lastly, we discuss some ethical issues regarding the sharing of protocols, training datasets, and HTR models on the Transkribus platform and between different platforms.

## II THE PROJECT AND ITS CORPUS

*Donner le goût de l'archive à l'ère numérique* is a broad partnership between the Université de Montréal, GLAM (Galleries, Libraries, Archives and Museums) and digital humanities research institutions. Led by Dominique Deslandres, the project aims to discover and make known the past of the Indigenous and Euro-Canadian communities that "made" Montréal in the 17th century. To do so, all collaborators adopted an intersectoral approach to the judicial and notarial archives produced in Montreal during this period to facilitate historical research, interpretation, collaborative transcription and indexing of the corpus. Overall, more than 50 people have been involved in this project.[4]

The documents examined for this project are preserved at Bibliothèque et Archive nationales du Québec (BAnQ) and were either produced by the *bailliage de Montréal*, a seigniorial court of justice that exercised its jurisdiction on the island of Montreal from 1640 to 1698, or by different notaries between 1648 and 1730. This set of documents can be divided into three main corpora totaling over 83,000 pages. 1- **The court records,** which consist of 35 volumes (about 5,000 pages) of proceedings, written by twelve clerks between 1665 and 1698. 2- The **files of the bailliage,** which are composed of 20 boxes (about 8,000 pages) of judicial documents written by more than twenty different people (among which we find several of the clerks identified in the registers) between 1644 and 1698. 3- The **notarial records**, that is some 70,000 pages of notarial acts produced by the fifteen notaries (some of them also acted as clerks for the court) active in Montreal between 1648 and 1730.

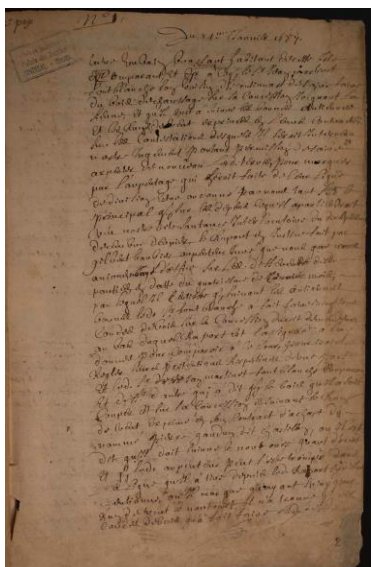 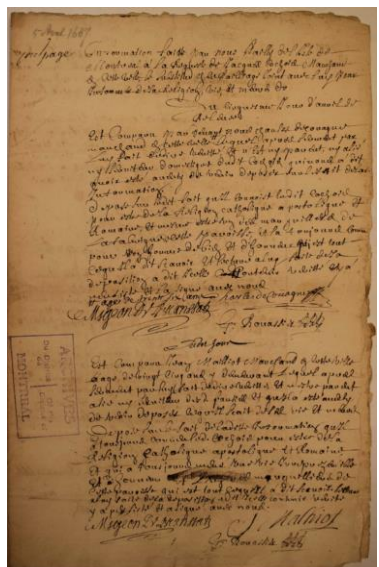 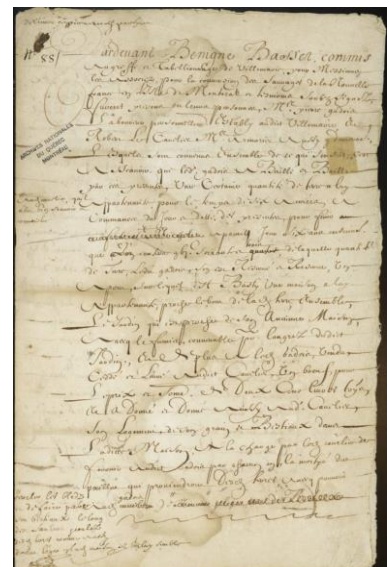

*Figure 1*. Page from the court records (BAnQ-Mtl, TL2, S11, D4, 11590 [1687], register 28, p. 3)

*Figure 2*. Page from the files of the bailliage (BAnQ-Mtl, TL2, 11571, p. 84)

*Figure 3*. Page from the notarial records of Bénigne Basset (BAnQ-Mtl, CN601, S17, p. 2)

To deal with these three bodies of documents, fifteen models were trained (see table 2). A first set of models was specifically produced for a single-hand use. Among these models, a particular attention was given to the one trained to process Claude Maugue's (1642-1696), Bénigne Basset's (1657-1699) and Antoine Adhémar's (1639-1714) handwritings, since more than 75% of the corpora is written by them (see table 1). Afterwards, we decided to invest time in the creation of general models which could be applied to pages written by multiple hands (in the files), as well as by notaries and clerks who produced a smaller proportion of the corpus. To do so, we first created single-hand models for clerks and notaries as illustrated in table 2.

---

[4] For the complete list of collaborators, whether institutional or personal contributors, see the project website (https://donner-le-gout-de-larchive.weebly.com/notre-eacutequipe.html).

|  | **Court records** | **Files**[5] | **Notarial records** | **Total** | **Total (%)** (rounded off to the nearest unit) |
|---|---:|---:|---:|---:|---:|
| **Adhémar** | 1,900 | 2,500 | 31,620 | 36,020 | 43 |
| **Bailly** | 5 | 250 |  | 255 | < 1 |
| **Basset** | 430 | 1,500 | 15,250 | 17,180 | 21 |
| **Bouassier** | 115 | 200 |  | 315 | < 1 |
| **Bourdon** |  |  | 90 | 90 | < 1 |
| **Bourgine** | 345 |  | 1,410 | 1,755 | 2 |
| **Bourgonnière** | 245 |  |  | 245 | < 1 |
| **Branssat** |  | 250 |  | 250 | < 1 |
| **Cabazier** | 120 | 200 | 1,020 | 1,340 | 2 |
| **Closse** |  |  | 200 | 200 | < 1 |
| **Delamaze** | 90 |  |  | 90 | < 1 |
| **Fleuricourt** |  |  | 300 | 300 | < 1 |
| **Laschney** |  |  | 1,170 | 1,170 | 1 |
| **Maugue** | 1,200 |  | 13,050 | 14,250 | 17 |
| **Saint-Onge** |  |  | 550 | 550 | < 1 |
| **Moreau** |  |  | 1,840 | 1,840 | 2 |
| **Mouchy** | 40 |  | 100 | 140 | < 1 |
| **Pottier** |  |  | 1,780 | 1,780 | 2 |
| **Pruneau** | 400 |  |  | 400 | < 1 |
| **Quesneville** | 10 | 200 |  | 210 | < 1 |
| **Raimbault** |  |  | 1,570 | 1,570 | 2 |
| **Saint-Père** |  |  | 200 | 200 | < 1 |
| **Unknown** | 105 | 2,900 |  | 3,005 | 4 |
| **Total** | 5,005 | 8,000 | 70,150 | 83,155 |  |

*Table 1*. Distribution of hands per page in the corpus

|  | **Pages produced** | **Words used for training** | **CER on validation set (%)** |
|---|---:|---:|---:|
| **Adhémar** | 36,020 | 22,639 | 8.40 |
| **Basset** | 17,180 | 6,407 | 9.90 |
| **Bouassier** | 315 | 3,847 | 16.80 |
| **Bourgine** | 1,755 | 9,536 | 14.81 |
| **Bourgonnière** | 245 | 5,829 | 11.70 |
| **Cabazier** | 1,340 | 19,196 | 16.80 |
| **Maugue** | 14,250 | 36,510 | 11.34 |
| **Mouchy** | 40 | 4,911 | 17.10 |
| **Saint-Onge** | 550 | 13,149 | 8.30 |
| **Pruneau** | 400 | 8,846 | 9.00 |
| **Bailliage of Montréal** | - | 73,720 | 10.12 |
| **Clerks of the bailliage** | - | 95,256 | 12.50 |
| **Canadian Notaries** | - | 225,919 | 10.70 |

*Table 2*. Final version of models we trained

In total, we estimate that more than 2,500 hours were invested in manually transcribing approximately 1,500 pages and cross-checking them as a group to create and evaluate models in a two-year period. Initially, a team of five people (Béatrice Couture, Farah Verret, Dominique Deslandres, Normand Robert and Maxime Gohier) took on the entire task, but due to the scale of the project, a call to the community was made. Since October 2021, a team of experienced paleographers has joined the project on a voluntary basis to participate in the verification of transcriptions and the production of models for notarial records during the Ateliers permanents d'analyse documentaire (APAD).

---

[5] Because of its heterogeneous nature (several hands on the same page), a complete inventory of the hands has not been made and the numbers presented are a gross aproximation.

## III METHODS

When we started to produce HTR models, our team decided to focus on single-hand models for clerks who worked at the *bailliage* and wrote at least a hundred pages (see table 1). Therefore, we carried out a systematic inventory of the hands present in the court records. As illustrated in table 1, Maugue and Adhémar wrote respectively 24% and 38% of the court records' pages and were prominent notaries, as they produced 2,175 and 5,000 notarial acts. This justified the training of different iterations for both models, where we compared different parameters in order to obtain satisfactory results (less than 10% CER).

Alongside the creation of the "Maugue" and "Adhemar" models, models specific to the handwriting of six other clerks appearing in the court records were trained during the summer and fall of 2021 (see table 2). However, since these clerks wrote on average only 250 pages each in this corpus, it did not seem necessary to invest too much time to achieve high performance. These models were primarily intended to create a preliminary transcription that would then be corrected manually to retrieve prosopographical information. Therefore, we have trained models for all the clerks who wrote at least a hundred pages in the court records and, to do so, we manually transcribed about fifty pages for each of these hands. The CER on validation data sets varied between 9% and 17%, a variation mainly related to the quality and regularity of the writing. The APAD team started to manually transcribe the acts of different notaries (Saint-Onge, Saint-Père, Mouchy, Closse, Laschney, Basset) and creating models. In these cases, we achieved CER on validation sets that varied between 8% and 17%.

By combining pages from all these writers, we have created a few iterations of three general models. The most advanced and sophisticated of them is called "Canadian Notaries – 17th century". This process allowed us to evaluate, in the long run, the influence of several factors on model training. Before introducing these factors, we must address the issue of performance computation and the context in which the models were created.

### 3.1 Performance computation

Evaluating the performance of a text recognition model is essential to the use of HTR technologies. By doing so, one not only avoids unnecessary work, but also limits the costs associated with automated transcription. Such an evaluation is based on various statistical data, including the Character Error Rate (CER) (READ-COOP-b, 2021) and Word Error Rate (WER). These figures indicate the number of characters or words incorrectly transcribed by the algorithm. In general, the CER is the easiest, most common and useful measure to assess a model's performance and compare it with others, even if other measures can be used, such as the bag-of-words F1-measure (Ströbel et. al, 2020).

In order to measure more precisely the performance of a model on specific documents, it is possible to systematically compare the reference transcription (the verified one) of any page with the inference (called "hypothesis" in the Transkribus interface and prediction in other platforms) proposed by automatic transcription of the same page with an HTR model. The result obtained from such a comparison is the most meaningful, since the different models can be compared with a control sample.

## 3.2 The creation of models

*3.3.1 Model for Claude Maugue*
In the summer of 2021, the model "Maugue" was trained to transcribe the 1,200 pages of the court records written by Claude Maugue. The goal was not to provide a perfect transcription of this corpus (which, incidentally, is almost impossible), but only to facilitate the retrieval of prosopographic information concerning Montreal's population. Creating an accurate model was, in Maugue's case, a real challenge for several reasons (see figures 4 and 5): firstly, the quality of his writing varied considerably throughout the corpus (due to his alcoholism); secondly, many documents showed ink smudges and bleed-through; thirdly, the corpus was composed of both high-quality color photographs and ordinary microfilm scans, some of which were from positive reels (black writing on a light background) while others were from negatives (white writing on a dark background). It was thus decided to manually transcribe 154 selected pages (totaling approximately 36,500 words) written by this clerk, including 97 pages from the court records and 57 pages from the files and notarial records, a selection that illustrates his erratic and often messy handwriting. It was also deemed necessary to incorporate as many instances as possible of elements typical of his writing habits, such as special abbreviation marks, ascending and descending stems extending into the top and bottom lines or smudges due to ink overuse. In addition, since the Transkribus algorithm struggles to deal with reverse contrasts, we removed from our training set 8 of the 154 pages, as these were from negative microfilm. It is in fact impossible to run HTR models on reverse contrast (even layout analysis) without first inverting colors with a special software.

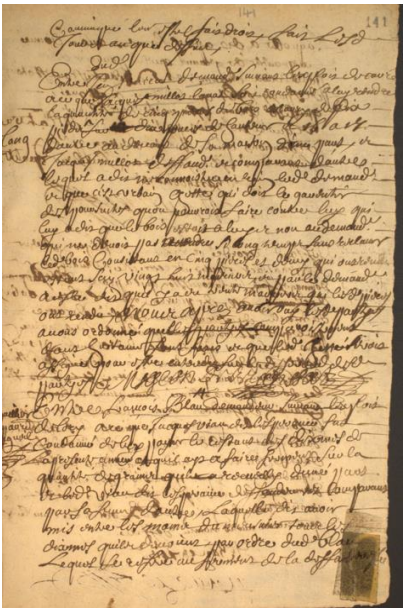 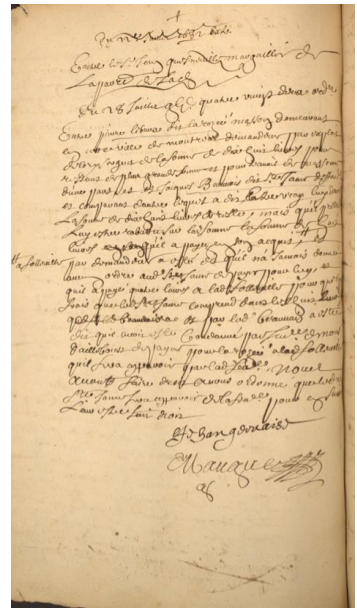

*Figure 4.* **Example of a page written by Maugue that is difficult to decipher**     *Figure 5.* **Example of a well-written page by Maugue**

After training over 400 epochs, this model achieved a CER of 11.34% on validation set. However, comparative tests were conducted on a sample of 10 selected representative pages, which revealed an average CER of 5.88%, a result quite satisfactory for our needs.

*3.3.2 Model for Antoine Adhémar dit Saint-Martin*
In the fall of 2021, the team set about to create the model "Adhemar". The experience acquired previously proved very useful when it came to selecting the pages to transcribe. Moreover, the transcription of the sample was quite easy since Antoine Adhemar was one of New France's most prolific notaries. Furthermore, he had a very regular handwriting which, although raising

Journal of Data Mining and Digital Humanities                                                          http://jdmdh.episciences.org
ISSN 2416-5999, an open-access journal

some deciphering challenges, is well known to experts, many of whom voluntarily helped transcribe our sample. Therefore, 133 pages were transcribed from the court records (22,639 words). After training for 400 epochs, the model achieved a CER of 8.15% on the validation data. As we conducted comparative tests on a sample of 10 selected representative pages, we achieved an average CER of 5.04%. This result means that the Adhemar model outperform Maugue's model for their own respective set of pages.

*3.3.3 Towards a successful general model: Canadian Notaries – 17th century*
After the creation of more than eleven single-hand models, we decided to combine the transcription of 687 pages from the court records to 326 pages of notarial acts, which were produced by nine hands (see table 3). At this point, we had a total of 1013 pages of ground truth material that we used to train a universal model combining two very different corpora. After training for 400 epochs, "Canadian Notaries – 17th century" achieved a CER of 10.70% on the validation set, while the comparison tests on 10 new pages selected in the files from the *bailliage* (from six different hands) revealed a CER of 7.17%. Thanks to its robustness and versatility, this model has allowed us to efficiently complete the transcription of all the Montreal bailliage's documentation (records and files representing more than 10,000 pages) and shows great promise for continuing the transcription of the notarial records.

| **Model** | **Words used for training** | **CER on validation set (%)** |
|---|---:|---:|
| **Adhémar** | 22,639 | 8.40 |
| **Basset** | 6,407 | 9.90 |
| **Bourgine** | 3,847 | 14.81 |
| **Bourgonnière** | 6,330 | 11.70 |
| **Maugue** | 36,510 | 11.34 |
| **Saint-Onge** | 13,149 | 8.30 |
| **Pruneau** | 8,846 | 9.00 |
| **Bailliage of Montréal** | 73,720 | 10.12 |

*Table 3*. Individual datasets combined to create *Canadian Notaries – 17th century.*

**IV RESULTS AND DISCUSSION**
The experience gained from training all these models enabled us to make some observations about the elements influencing the performance of an HTR model. These findings are perhaps not based on an advanced knowledge of AI algorithm functioning and, as such, might seem somewhat obvious to computer science and deep learning specialists. Nevertheless, they are grounded on empirical evidence gained from extensive use of the platform, rather than predictions made based on software programming choices or its source code structure. Of the set of elements that can affect the performance of HTR models, three appeared especially significant to us: the transcription protocol, the language model functions associated with the training data and, finally, the use of pre-existing base models.

**4.1 The importance of a transcription protocol**
Transkribus offers great opportunities for collaborative work, and we seized this opportunity to create a team of five paleographers (which quickly grew to fourteen) who started to work on several pages of transcription. However, because each transcriber of the team was trained differently in paleography, the transcriptions produced were not uniform. This peculiarity led to disappointing results with the first iteration of the "Maugue" model, especially regarding abbreviations. Indeed, HTR algorithms can handle abbreviations either by transcribing only the abbreviated form present in the text or by expanding them with the missing characters (Thöle, 2017; Stutzmann, 2017). The solution chosen depends on the data provided in the training process. But if both systems (abbreviated and expanded forms) are integrated within the training data, the algorithm will be confused as to which option to choose and might produce many errors. This was a major problem with our corpora (which contain a lot of abbreviations), and the poor performance of this first model of Maugue was a direct result of the discrepancy in transcribers' paleographic practices. Thus, like other researchers (Pinche, 2022), we realized the fundamental need to have a common transcription protocol and the need to train all transcribers in its rigorous application. We even had to go back to our transcription to apply these guidelines to previously transcribed pages. Of course, in paleography, as well as in machine learning, this reality is fundamental and always represents an important issue as there is no universal approach to manuscript transcription.

To make sure that the training data would be as exact as possible, we decided to create a rigorous protocol as well as a procedure for validating transcriptions and address paleographic issues.

Inspired by the protocol developed by Maxime Gohier for the project *Nouvelle-France Numérique*[6], we chose to rigorously replicate the spelling of every word, and also the presence or absence of punctuation. It was thus decided not to standardize individuals' first and last names, nor location names or common words. However, not being yet familiar with the methodological concepts of graphemic and graphetic transcriptions (Stutzmann, 2010), we had not expected some issues such as the varying ways in which individual scripters sometimes write abbreviations or use particular glyphs or symbols. The most evocative example of this problem concerns the word *ledit*, generally abbreviated as "led.", and all its derived forms (*ladite* = lad.; *lesdits* = lesd.; *audit* = aud.; *auxdits* = auxd.; etc.). Probably the most common abbreviation in court and notarial documents, it is written in various ways. Some clerks, like Maugue, use a final plunging character resembling a "y" as an abbreviation mark, while others, like Adhémar, simply make a common "d." or extend the final stroke through the top of the line. As we were creating general models, it became clear that a graphetic transcription would only create confusion for the algorithm. Therefore, in order to homogenize transcriptions, it was decided to follow a rigorous graphemic methodology and to consider the final letter used by Maugue as a "d" regardless of its plunging form. This kind of decision was also taken for other cases, such as the Latin ligature *ou* (rendered "8" instead of "ȣ") and the different monetary signs ("₶" for *livre*, "s." for *sol* and "d." for *denier*).

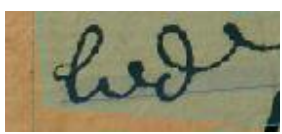 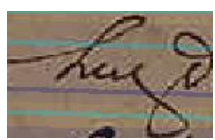 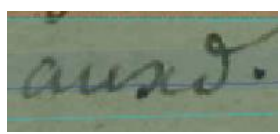 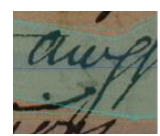

*Figure 6.* **Abbreviation of ledit.**    *Figure 7.* **Abbreviation of ledit.**    *Figure 8.* **Abbreviation of auxdits.**    *Figure 9.* **Abbreviation of auxdits.**

On the other hand, we decided to organize weekly working sessions to counter-verify the transcription, identify unclear words and discuss paleographic issues. This allowed us to get higher quality data and enlarge our training dataset. Indeed, when a word is tagged <unclear> in Transkribus, the model training algorithm excludes not only the single word, but the whole line where this word appears (READ-COOP-a, 2022). Each <unclear> word deciphered might thus allow retrieving 5 to 10 words from the dataset. In the end, this seemingly time-consuming procedure has proven highly effective.

The implementation of these rules had a significant impact on the quality of our HTR models. Comparison tests on the same set of pages using two versions of the "Maugue" model showed a clear improvement in the CER on the validation data, which went from 14.10% to 11.34%, a decrease of 3 points or a proportional decrease of about 19.57% of the errors.

---

[6] This protocol is under development and should be published soon on the website (https://nouvellefrancenumerique.info).

Since the beginning of our work, a few transcription protocols have been proposed by organizations or research teams in order to guide the users of HTR and Transkribus (notably those of READ-COOP[7] and CREMMALab[8]). Although they do not cover all issues raised by the conversion of handwriting into machine-readable data (which, in any case, is almost impossible), these protocols can serve as guidelines for the development of custom rules by suggesting a general approach to address very specific problems raised by each corpus as they provide insight on the way AI processes these problems according to the parameters of HTR algorithms. These protocols can indeed help, for example, identify more clearly the specific needs of a project regarding automated transcription, or determine which types of data deserve to be standardized and which do not need to be, or should not be.

It is therefore essential to take the time to properly plan the initial transcription work, as this step is critical in the production of a model's training data. Such an investment really pays off in the long run because, in the end, the potential for, and accuracy of, analysis of the documentation will be linked to the quality of transcriptions generated by the model.

### 4.2 The impact of Language Models

HTR technologies are much more efficient at processing handwritten material than traditional OCR for several reasons. Of course, the main reason is associated with the fact that HTR models can learn to recognize an almost infinite number of shapes for the same character or glyph. A second reason is the fact that HTR technologies can incorporate statistical Language Models (LMs) to enhance their efficiency. Instead of recognizing each character individually, HTR algorithms such as PyLaia use *n*-gram models to generate character strings, thus having a higher probability of corresponding with the training data (De Mulder et. al., 2014; Tassopoulou et al., 2020). This represented a novelty when HTR technologies were developed, and OCR technologies had yet to implement this functionality[9], as they were previously based on predetermined lexica or dictionaries (Smith, 2011). Therefore, the PyLaia technology can "learn" to recognize strings of characters that form words or sub-words specific to the corpus.

The Transkribus platform allows users to choose whether to use this feature when a text recognition is initiated. If chosen, during the recognition process, the model will use a LMs created from the training data to infer a string of characters. According to [Günter Mülhberger in a non-published communication, 2022[10]], the use of such a feature could reduce erroneous words (WER) by up to 10%. Although we have never achieved such results, it seems clear that this feature almost systematically has a significant impact on error rates. However, to take full advantage of it, it is important to understand how LMs interfere in the model's training and recognition processes.

Our experience has shown two methods for taking advantage of a language model. The first is to vary the vocabulary embedded in the training data as much as possible. To do so, the pages to be transcribed should be selected so that they represent as closely as possible the entire corpus. If the corpus extends over several years, this method will have the advantage of considering not only the evolution of the writer's writing style, but also the variety of vocabulary contained in the different types of documents. In the case of judicial documents, it seems preferable to select a few pages from several trials rather than completely transcribe a few of them. Within a single trial, the lexicon is generally quite homogeneous, both in terms of technical legal terms and names of individuals and places involved.

---

[7] Transkribus Transcription Convention, READ-COOP. (https://readcoop.eu/transkribus/howto/transkribus-transcription-conventions/)

[8] *Guide de transcription pour les manuscrits du Xe au XVe siècle*, Centre Jean Mabillon. (https://hal.science/hal-03697382)

[9] Language models are more and more used by OCR developers. Tesseract is probably one of the first to have integrated this technology to its core. (Dobrac: https://link.springer.com/article/10.1007/s10032-020-00359-9)

[10] École d'été en culture numérique et gestion des données de recherche, Université du Québec à Rimouski, 2022

Another method consists in exploiting the indexes that can be found in an archive file. Indexes are very useful for training HTR models since they usually provide onomastic information representative of the entire file they describe. In the court records of Montreal's bailliage, we were lucky enough to discover an index of 42 pages drawn up by Adhémar, in which he listed the plaintiffs and defendents of all the cases he handled. The addition of these pages to the training data of the "Adhemar" model had a rather small impact on the CER, but a significant one on its ability to recognize proper names. As we trained three iterations of the model[11] and compared them on the same five pages, we saw that the model with the index had the best CER (5.64%) while the highest CER was from the model without the index (6.31%). Though this may seem nonsignificant, we saw that there were more errors in onomastic information for the iteration without the index while the other two iterations presented fewer. The inclusion of Adhemar's index was therefore useful in view of the project's main objective, that is identifying 17th-century-Montreal's population in the archives. Thus, the sampling of data to be transcribed appears to be a strategic step in the process of training HTR models, a step that requires a certain investment of time to ensure that the data selected is representative of the entire corpus to be processed and that it is consistent with research objectives.

### 4.3 The fundamental role of base models

The last element we have worked on to improve the performance of our models is related to the usage of base models. Base models are an optional feature of an HTR training algorithm that uses the fine-tuning techniques of transferring information from a pre-existing model into the newly trained model. By doing so, the training algorithm utilizes knowledge acquired by a previous model (usually based on a big dataset) to pre-calibrate its processes of character extraction (image segmentation) and classification of extracted characters. This allows to create a performant model even with a dataset that is quite small, or to speed up the training of a model by reducing the number of epochs required. More precisely, as [Ströbel et al, 2022] specify, "A base model initializes the model's weights and allows for fine-tuning the model on novel data. This way, the model knows something about handwriting before seeing the new training data, leading to faster convergence and better performance."

During our project, we tried to measure empirically the impact of base models on the training of a new model, more specifically focusing on how to choose the best base model. First and foremost, we found that the use of a base model can contribute to reducing the CER on the validation set of a model by up to 66%. Indeed, the "Adhemar" model without a base model gave a CER on the validation data of 25.40%, while the same model trained with the base model "Bailliage of Montreal" reaches a CER of 8.40% on the validation set. This base model is trained on approximately 300 pages (73,720 words) from the bailliage and achieved a CER on validation set of 10.12%. As for Maugue's model, it reached a CER of 18.60% on the validation set without the base model, while with the base model "Baillage of Montreal", it reached 11.34%. The usefulness of such a feature should therefore not be underestimated, and Transkribus users should take advantage of this optional feature, especially when training small models.

---

[11] 1. Adhemar with index (22,639 words); 2. Adhemar without index (17,117 words); 3. A smaller version of Adhemar with index (16,370 words).

What is even more remarkable in these results, however, is the fact that this base model (Bailliage of Montreal) offered the largest decrease in CER, even though we created iterations with base models from the same period that performed better and had considerably more training data. Indeed, we used the "New France 17th-18th Century" model, which is much larger (304,325 words) and theoretically a lot better performing (with a CER of 5.00% on its validation set), as a base model for Maugue's models. The "New France 17th-18th Century" model is based on a corpus written mainly between 1660 and 1715 by secretaries of governors and intendants of New France. Maugue's model achieved a higher CER during tests (7.87% for New France and 5.88% for Bailliage of Montreal), which means an increase in errors of over 2% (or a proportionate increase of 25.28% of the CER). This drop in performance is clearly a result of the different lexica used in the base model and in the documents written by Maugue, as the bailliage records contain very specific and redundant legal jargon, while the elite's correspondence deals with a wider variety of topics using much simpler vocabulary.

|  | Number of words | CER on validation set (%) | CER of Maugue's model with base model (%) |
|---|---|---|---|
| **New France 17th-18th c.** | 304,325 | 5.00 | 10.12 |
| **Bailliage of Montreal** | 73,720 | 7.87 | 5.88 |

*Table 4.* Performance of different base models used for training *Maugue's* model.

These observations demonstrate that the use of a base model is always helpful. Furthermore, to significantly increase the performance of a new model, the base model used does not need to rely on data similar to the newly trained models. Also, the size of the base model dataset is not necessarily a guarantee of quality for the resulting model, and to create an efficient model, it seems preferable, for a corpus written by a homogeneous hand, to use a more specific base model that contains data similar (especially from a lexical point of view) to the one present in the corpus of the new model. A base model's similarity will then have a greater impact than its size or computed CER performance.

## V CONCLUSION

The experience acquired from the project *Donner le goût de l'archive* highlights the relevance of understanding the impacts of several variables on HTR algorithms in order to increase the quality of recognition models. In this perspective, we have shown how the development of a rigorous transcription protocol appears to be a critical step, since the performance of a model depends primarily on the quality of its training data. It is essential to ensure the homogeneity of this data, and failing to do so presents a risk of misleading the model, especially with abbreviations. It also seems important to carefully select the training data, to make sure that it is as representative of the corpus to be transcribed as possible, not only in terms of spelling, but also (and especially) in terms of vocabulary (the lexicon). A detailed knowledge of the corpus to be transcribed is always helpful in determining the preferred sampling method. Moreover, a good understanding of the way lexica influence the recognition process helps increase the profitability of the time and resources invested in the preparation of training data. Finally, while it is obvious that the use of base models for training a new model is advantageous, it is also relevant to consider the nature of the data on which these base models were trained. However, on this point, we think it is relevant to emphasize that a clearer view from developers regarding the processes through which base models are exploited in their engines would greatly help users better understand, and benefit from, their uses.

This said, our experience in creating HTR models has also raised certain ethical issues that we feel are relevant to share here, in order to increase the profitability of the efforts invested by the community in the training of HTR models, a time-consuming process. Failing to question our use of this technology, we risk missing an extraordinary opportunity to put collective work at the service of democratization of research in (and on) manuscript documents.

One of the major challenges we are facing regards the sharing of models and training data. As a cooperative, the READ-COOP and its Transkribus platform aim to facilitate the sharing of this data, by allowing users to make their models "public" (which means within the Transkribus community). However, despite the ease provided by the platform to do so, few people bother to share their data. We ourselves have not made public any of the models or datasets produced in *Donner le goût de l'archive*, despite the fifteen or so models trained, and the thousands of pages of documents transcribed. Three major reasons explain this: first, the lack of time; second, the fact that none of our models seemed good enough to be made public yet; and third, the obligation to obtain authorization from the institutions that own the archival images. Our case is far from being exceptional, and instead seems to be the norm. Indeed, the Transkribus website informs us that to date, more than 12,000 recognition models have been trained by the 100,000 or so users of the platform. Even if only a small fraction of these 12,000 models is really useful (the majority being merely tests in development processes), the fact remains that no more than 93 models are currently accessible to the whole community and sufficiently documented for users to judge their usefulness (READ-COOP-b, 2022). As all these models remain "private", the data on which they are based also remains inaccessible to most users, who are forced to start their own training data production without a base model or lexicon. It is true that the Transkribus development team has already produced a few meta-models that are accessible to all users, but these can still only be used in a few languages and are far from covering all historical periods. In addition, some initiatives have recently emerged to facilitate the sharing of "ground truth" data by users of any HTR tool, in order to make the training of models easier. The HTR-United platform, for example, launched in 2021 by Alix Chagué and Thibault Clérice, aims to facilitate the sharing and retrieval of useful datasets for model training. This platform even promotes the use of Creative Commons licenses to ensure the recognition of copyrights on the datasets thus made public. So far, however, the contribution of Transkribus users to this platform remains quite small (about 20 individuals) compared to that of eScriptorium users. We note that our enthusiasm to take advantage of HTR as quickly as possible to transcribe ever larger corpora with ever-lower error rates tends to lead us to lose sight of the essential element of artificial intelligence, namely the sharing of basic data. Indeed, this sharing is key if we want to maximize the profitability of the work done.

The open sharing of training datasets, which include document images and transcriptions, raises another ethical issue: that of data quality control. Currently, there is no way to verify the quality of datasets or models made publicly available, either directly in Transkribus or on platforms such as HTR-United. If these datasets presented as "ground truth" contain transcription errors or are associated with poor quality images, potential users have few ways of knowing this until they have tested them themselves; they must trust the data producers' competence and good faith. In Transkribus, for example, the model training algorithm can remove lines from the training data where text has been tagged <unclear>, to avoid inducing errors. Nevertheless, if this dataset is made public afterwards, it will retain the potentially erroneous data (all text tagged <unclear>), which can sometimes be quite considerable.

From all this, the question arises as to whether it would be necessary and relevant to establish validation bodies for the data produced for and by HTR. This question amounts to asking if the simple production of data should not be subjected to a peer review procedure before being made public, as is the case for all so-called scholarly publications. Of course, this would be a considerable challenge given the paleographic and linguistic skills needed to adequately assess the quality of a training dataset, not to mention the burden this would put on already overstretched researchers. Should we instead appeal to the principles of citizen science, as Wikipedia, for example, does by allowing readers the opportunity (and the burden!) to highlight or correct any information deemed false or irrelevant? In a way, doesn't the obligation to rely on the good faith or reputation of the data's producer contradict the very foundation of science, which postulates that truth can and must be deduced by reason?